\documentclass[letterpaper]{article} % DO NOT CHANGE THIS
\usepackage{aaai2027}  % DO NOT CHANGE THIS
\nocopyright
\usepackage[hyphens]{url}  % DO NOT CHANGE THIS
\usepackage{graphicx} % DO NOT CHANGE THIS
\usepackage{natbib}  % DO NOT CHANGE THIS AND DO NOT ADD ANY OPTIONS TO IT
\usepackage{caption} % DO NOT CHANGE THIS AND DO NOT ADD ANY OPTIONS TO IT
\usepackage{algorithm}
\usepackage{algorithmic}
\usepackage[utf8]{inputenc}
\usepackage[T1]{fontenc}
\usepackage{textcomp}
\usepackage{makecell}   % 允许表头换行
\usepackage{siunitx}    % 对齐百分比数值（可选）
\usepackage{subcaption}
\usepackage{amssymb}
\usepackage{multirow}
\usepackage{colortbl}

\usepackage{newfloat}
\usepackage{listings}
\DeclareCaptionStyle{ruled}{labelfont=normalfont,labelsep=colon,strut=off} % DO NOT CHANGE THIS
\floatstyle{ruled}
\newfloat{listing}{tb}{lst}{}
\floatname{listing}{Listing}

\usepackage{booktabs}

\title{MechReason: Benchmarking Multi-Image Multi-Hop Reasoning in Mechanical Engineering}
\author{
    Tengyue Wang\textsuperscript{\rm 2}\equalcontrib,
    Kang An\textsuperscript{\rm 1}\equalcontrib\thanks{Project leader},
    Chenxu Du\textsuperscript{\rm 5},
    Zhongyu Yang\textsuperscript{\rm 3},
    Yuanchi Zhu\textsuperscript{\rm 6,\rm 7},\\
    Xinqi Yang\textsuperscript{\rm 8},
    Hebao Zhu\textsuperscript{\rm 9},
    Ziliang Wang\textsuperscript{\rm 4},
    Faqiang Qian\textsuperscript{\rm 4},
    Yunli Yang\textsuperscript{\rm 10},
    Qibing Ren\textsuperscript{\rm 1}\corresponding
}
\affiliations{
    \textsuperscript{\rm 1}Shanghai Jiao Tong University,
    \textsuperscript{\rm 2}South China University of Technology,
    \textsuperscript{\rm 3}ModelBest,
    \textsuperscript{\rm 4}SenseTime,\\
    \textsuperscript{\rm 5}Southwest Jiaotong University,
    \textsuperscript{\rm 6}ShanghaiTech University,
    \textsuperscript{\rm 7}Institute of Automation, Chinese Academy of Sciences,\\
    \textsuperscript{\rm 8}East China Normal University,
    \textsuperscript{\rm 9}Chongqing University,
    \textsuperscript{\rm 10}Institute for Advanced Algorithms Research, Shanghai\\
    tengyuew32@gmail.com,
    \{an\_kang, renqibing\}@sjtu.edu.cn
}

\begin{document}

\maketitle

\begin{abstract}
Despite significant progress in general visual question answering and cross-modal understanding, multimodal large language models still face a pronounced gap in evaluation for complex reasoning within the mechanical engineering domain. Existing benchmarks predominantly focus on rudimentary tasks such as drawing recognition, CAD interpretation, or single-chart querying, falling short of assessing whether models can integrate multiple images, textual conditions, physical principles, and engineering constraints to perform multi-step reasoning when confronted with authentic, intricate mechanical problems. To address this, we introduce MechReason, a benchmark derived from real mechanical engineering papers, comprising 12k question-answer pairs with explicit reasoning-chain annotations and 21k visual materials spanning nine evidence types, including statistical charts, parameter tables, engineering drawings, microscopic images, simulation images, system architectures, real mechanical scene photos, CAD model images and manufacturing flowcharts. MechReason covers eight task types across four reasoning dimensions: explanation, prediction, design, and diagnosis. We devise a four-stage construction pipeline: we first extract core engineering claims and decompose their supporting evidence into premises, reasoning processes, conclusions, and corroborative evidence; we then generate shortcut-preventing questions by masking posterior verification information; finally, we apply multimodal quality validation to ensure task quality and multi-hop nature. Extensive experimental results demonstrate that MechReason is highly challenging, with even the most advanced models achieving only 62.89\% accuracy.
\end{abstract}

\begin{links}
    \link{Code}{https://github.com/Lifelong-journey/MechReason}
\end{links}

\section{Introduction}

Mechanical engineers often use experimental figures, simulation maps, or performance tables as posterior evidence to validate final conclusions. Although multimodal large language models (MLLMs) have made substantial progress in general visual question answering and scientific chart understanding \cite{masry2022chartqa, lu2022learn, yue2024mmmu, pramanick2024spiqa}, their ability to perform complex mechanical reasoning remains insufficiently evaluated. Existing mechanical and industrial benchmarks mainly focus on localized perception or shallow reasoning tasks, such as drawing interpretation, CAD understanding, rule lookup, or simple calculation \cite{doris2025designqa, kunz2025techmb, mallis2026text, doris2026cadbench, kou2026mechvqa}. As shown in Table~\ref{tab:benchmark_comparison}, they rarely combine heterogeneous visual evidence, multi-image inputs, complex engineering claims, and explicit reasoning supervision, leaving multi-hop mechanical reasoning largely unexplored.

\begin{figure*}[htbp]
  \centering
  \includegraphics[width=0.95\textwidth]{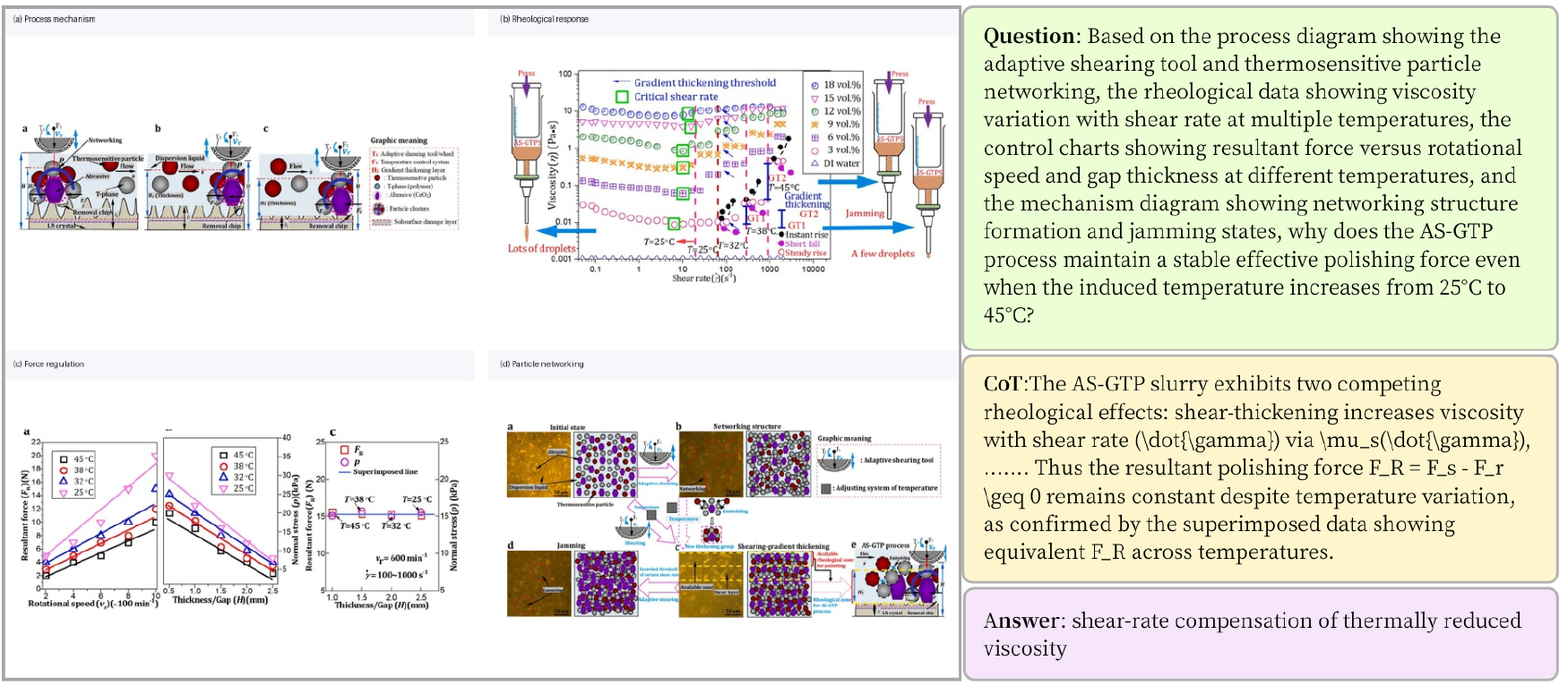}  
  \caption{A MechReason case requiring multi-image reasoning over process mechanism, rheology, force regulation, and particle-network evolution.}
  \label{fig:caseStudy}
\end{figure*}

In this work, we introduce \textbf{MechReason}, the first multi-image, multi-hop reasoning benchmark for mechanical engineering. We use scientific papers as data sources because they naturally contain complex engineering conclusions, explicit argumentation processes, and heterogeneous evidence close to real-world mechanical scenarios. MechReason contains 12,257 question-answer pairs, including 9,994 training samples and 2,263 test samples, with explicit chain-of-thought (CoT) supervision. It includes 21,842 unique visual materials spanning nine types of mechanical evidence. Questions are organized into four reasoning intentions: \textit{Explanation}, \textit{Prediction}, \textit{Design}, and \textit{Diagnosis}. Each question reflects an engineering-oriented task, requiring the model to synthesize structure, materials, operating conditions, physical mechanisms, and multi-source visual evidence to perform real-world scenario tasks such as design analysis, performance prediction, fault diagnosis, and process optimization. All questions and answers have been validated by mechanical engineers to ensure technical correctness and practical relevance. Figure~\ref{fig:caseStudy} shows a representative case where the model must integrate process mechanism, rheological response, force regulation, and particle-network evolution to infer the underlying mechanical conclusion.

Constructing MechReason is non-trivial. We observe that directly prompting models to generate questions from papers often causes them to include posterior verification evidence, such as final performance tables, simulation maps, or experimental figures, in the question context. Such evidence is originally used by authors to validate conclusions, but once exposed to the solver, it can turn a complex reasoning problem into simple figure reading, table lookup, or OCR. We therefore propose a four-stage construction pipeline: identifying core engineering claims, decomposing each supporting argument into prerequisites, reasoning process, conclusion, and verification evidence, generating anti-shortcut questions by exposing only necessary prerequisites while reserving the reasoning process as CoT supervision, and applying multimodal quality verification.

We evaluate representative proprietary and open-source MLLMs on MechReason. Even the best-performing model achieves only 62.89\% accuracy, showing that complex multi-image mechanical reasoning remains highly challenging. Fine-grained analysis reveals that errors mainly arise from flawed reasoning chains, evidence grounding failures, and lossy answer compression, rather than simple hallucination. We further show that CoT-based supervised fine-tuning improves performance, but substantial gaps remain.

Our main contributions are summarized as follows:
\begin{itemize}
    \item We present \textbf{MechReason}, the first multi-image, multi-hop reasoning benchmark for mechanical engineering, covering engineer-oriented tasks such as mechanism analysis, performance prediction, design selection, and fault diagnosis across heterogeneous visual evidence and real-world mechanical scenarios.
    \item We propose a four-stage construction pipeline that separates prerequisites, reasoning processes, conclusions, and verification evidence, reducing hallucination and pseudo multi-hop shortcuts.
    \item We systematically evaluate state-of-the-art MLLMs, analyze their failure modes in complex mechanical reasoning, and investigate the effect of CoT-based supervised fine-tuning.
\end{itemize}

\begin{table*}[t]
\centering
\footnotesize
\setlength{\tabcolsep}{4pt}
\renewcommand{\arraystretch}{1.0}
\caption{Comparison with representative mechanical and industrial multimodal benchmarks. 
``\checkmark'' denotes full support, ``$\triangle$'' denotes partial support, and ``--'' denotes not supported or not applicable.}
\label{tab:benchmark_comparison}
\begin{tabular}{lccccccc}
\toprule
Benchmark & Scale & Hetero. Visuals & Multi-Image & Multi-Hop & Paper Based & CoT Sup. & Features \\
\midrule
TechMB & 947 QA & -- & -- & $\triangle$ & -- & -- & Manufacturability \\
DesignQA & 1.45K QA & \checkmark & $\triangle$ & $\triangle$ & -- & \checkmark & Engineering Compliance \\
TriView2CAD & 203K samples & -- & \checkmark & $\triangle$ & -- & $\triangle$ & CAD Projection QA \\
BenchCAD & 17.9K parts & -- & \checkmark & $\triangle$ & -- & -- & CAD Code \\
CADBench & 18K samples & -- & $\triangle$ & -- & -- & -- & CAD generation \\
MechVQA & 21K QA & -- & $\triangle$ & $\triangle$ & -- & \checkmark & Engineering drawing \\
\rowcolor{gray!12}
MechReason & \textbf{12.3K QA} & \checkmark & \checkmark & \checkmark & \checkmark & \checkmark & Mechanical reasoning \\
\bottomrule
\end{tabular}
\end{table*}

\section{Related Work}
\subsection{LLMs for Mechanical Engineering}

In recent years, multimodal large language models (MLLMs) have been increasingly applied to mechanical engineering tasks, including design, manufacturing, and engineering analysis, driving the rapid development of corresponding evaluation benchmarks. However, existing studies primarily focus on isolated capabilities such as CAD modeling and engineering drawing understanding, while systematic evaluation of complex engineering reasoning remains limited \cite{baker2025large}.

Existing benchmarks can be broadly categorized into two groups. The first category focuses on evaluating CAD-related capabilities. DesignQA assesses a model's ability to extract engineering rules from CAD images and technical drawings \cite{doris2025designqa}, while CADBench, BenchCAD and TriView2CAD emphasize editable CAD program generation and industrial-standard part modeling \cite{doris2026cadbench, zhang2026benchcad, chun2025creft}. The second category targets the understanding of engineering drawings and technical documents. TechMB evaluates the capability of vision-language models in manufacturability analysis based on technical drawings \cite{kunz2025techmb}, and MechVQA introduces a comprehensive benchmark consisting of real-world mechanical parts and assembly drawings \cite{kou2026mechvqa}.

These benchmarks have established a solid foundation for evaluating MLLMs in mechanical engineering. Nevertheless, most existing benchmarks focus on a single information modality, local perception, or single-step reasoning, lacking benchmarks that require multi-hop reasoning across heterogeneous sources of engineering information.
\subsection{Multi-hop Reasoning Question Construction}

Multi-hop question answering has long been adopted to evaluate a model's ability to perform compositional reasoning by integrating evidence from multiple sources \cite{yang2018hotpotqa} \cite{welbl2018constructing} \cite{qi2021answering}. In heterogeneous data settings, HybridQA and TAT-QA combine tables with textual information \cite{chen2020hybridqa} \cite{zhu2021tat}, MultiModalQA further incorporates images \cite{talmor2021multimodalqa}, and DocHop-QA targets multi-document, multimodal scientific question answering \cite{park2025dochop}. While these benchmarks have substantially advanced multi-hop reasoning research, they are not specifically designed for mechanical engineering scenarios.

Meanwhile, existing multi-hop datasets commonly suffer from the problem of \emph{spurious multi-hop reasoning}, where models can correctly answer questions by exploiting lexical co-occurrence or local shortcuts instead of performing genuine compositional reasoning \cite{chen2019understanding}. Recent studies have attempted to mitigate shortcut exploitation through compositional dependency construction, filtering strategies, and bottom-up question generation \cite{yang2025large} \cite{qiu2025bmgq}. However, existing approaches overlook a shortcut inherent in the argumentative structure of scientific papers: the final conclusions are often directly verified by subsequent experimental figures and tables, allowing models to obtain the correct answers without reconstructing the underlying reasoning process. MechReason explicitly separates reasoning evidence from verification evidence, thereby constructing questions that genuinely require engineering reasoning.
\section{MechReason Benchmark}

\subsection{Benchmark Overview}

\begin{figure*}[t]
  \centering
  \begin{subfigure}{0.48\textwidth}
    \centering
    \includegraphics[width=\linewidth]{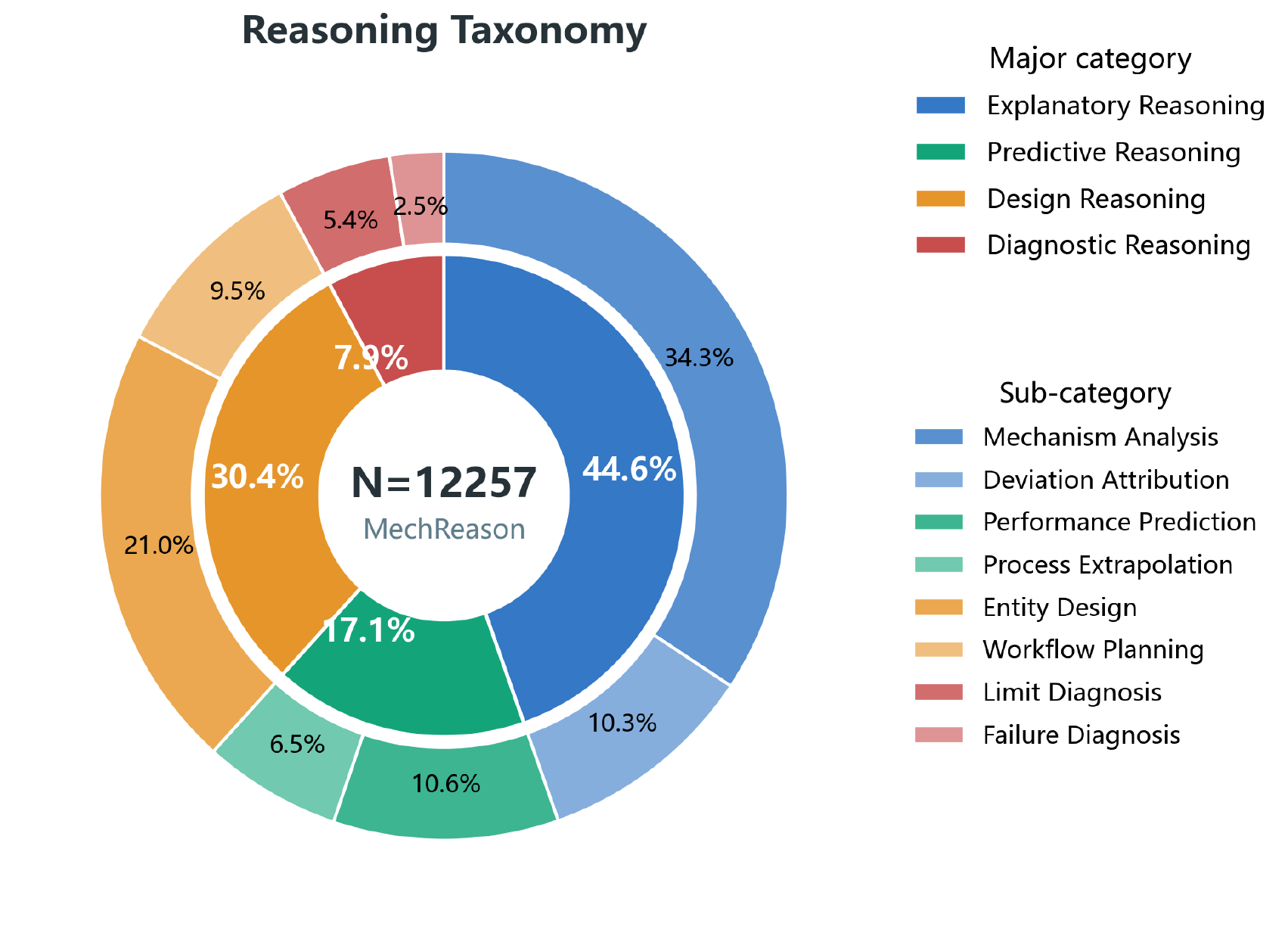}
    \caption{Reasoning taxonomy.}
  \end{subfigure}
  \hfill
  \begin{subfigure}{0.48\textwidth}
    \centering
    \includegraphics[width=\linewidth]{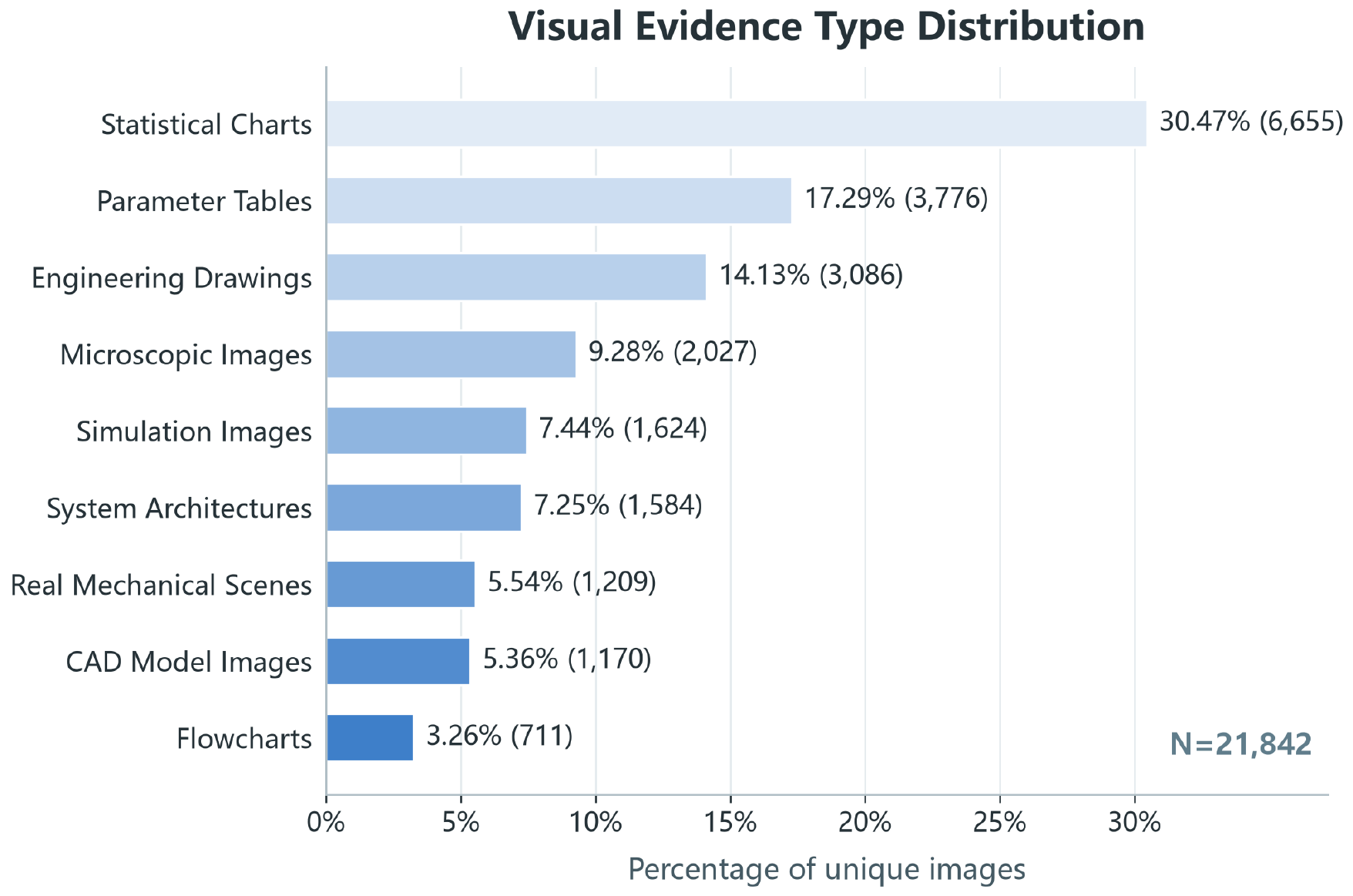}
    \caption{Visual evidence types.}
  \end{subfigure}
  \caption{Overview of MechReason. (a) Distribution of four reasoning dimensions and eight fine-grained subcategories. (b) Distribution of visual evidence types in MechReason.}
  \label{fig:Stats12}
\end{figure*}

MechReason is a multi-image, multi-hop reasoning benchmark for mechanical engineering. We use published papers as data sources because they naturally contain complex engineering conclusions, explicit argumentation processes, and heterogeneous evidence close to real mechanical scenarios. The benchmark covers representative mechanical engineering journals and spans thermal-fluid systems, manufacturing processes, machine tools, robotics, materials processing, structural analysis, and control systems.

The current version contains \textbf{9,994} training samples and \textbf{2,263} test samples, resulting in \textbf{12,257} QA samples in total. The training set is provided in two formats: direct-answer supervision and explicit reasoning-chain supervision. The test set is used for standard evaluation and includes annotations for question categories. Each sample consists of a question, a short answer, a reasoning supervision signal, and a set of relevant images. All test samples require visual information. Across all splits, MechReason contains \textbf{35,217} image references, with an average of \textbf{2.87} images per question. 

%\begin{figure}[t]
%  \centering
%  \includegraphics[width=\columnwidth]{Figures/visual_type.pdf}
%  \caption{Distribution of visual evidence types in MechReason.}
%  \label{fig:visualType}
%\end{figure}

MechReason contains \textbf{21,842} classified unique visual materials. As shown in Fig.~\ref{fig:Stats12}(b), these materials cover nine evidence types commonly used in mechanical engineering: statistical charts, parameter tables, engineering drawings, microscopic images, simulation images, system architectures, real mechanical scene photos, CAD model images and manufacturing flowcharts. This diversity distinguishes MechReason from benchmarks focused only on drawings or CAD models.

To characterize the reasoning abilities evaluated by MechReason, we construct a two-level taxonomy, as shown in Fig.~\ref{fig:Stats12}(a). Questions are first grouped into four reasoning dimensions according to the functional role of the answer: explanatory, predictive, design, and diagnostic reasoning. Each dimension is further divided into two subcategories.

\textbf{Explanatory reasoning} identifies the physical causes behind phenomena or performance conclusions. It includes \textbf{mechanism analysis}, which focuses on system-level causal mechanisms such as geometry, load transfer, multiphysics fields, and feedback interactions, and \textbf{deviation attribution}, which explains performance differences caused by material properties, process variations, measurement errors, or modeling simplifications.

\textbf{Predictive reasoning} infers system responses under changes in structures, operating conditions, or parameters. It includes \textbf{performance prediction} for comparing design alternatives, configuration variants, or material substitutions, and \textbf{process extrapolation} for predicting trends or sensitivities under parameter changes, state transitions, or boundary-condition perturbations.

\textbf{Design reasoning} selects feasible configurations, parameters, or strategies under engineering objectives and constraints. It includes \textbf{entity design}, concerning physical structures, key dimensions, allowable ranges, or manufacturing windows, and \textbf{workflow planning}, concerning control laws, modeling protocols, testing schemes, and optimization trade-offs.

\textbf{Diagnostic reasoning} traces the causes of performance limits, model breakdowns, or boundary violations. It includes \textbf{limit diagnosis}, which identifies physical bottlenecks such as heat-transfer limits, structural weak links, stability boundaries, or actuator limits, and \textbf{failure diagnosis}, which identifies failures caused by invalid assumptions, sensor noise, parameter coupling, or information degradation.

\subsection{Question Generation}

\begin{figure*}[t]
  \centering
  \includegraphics[width=0.9\textwidth]{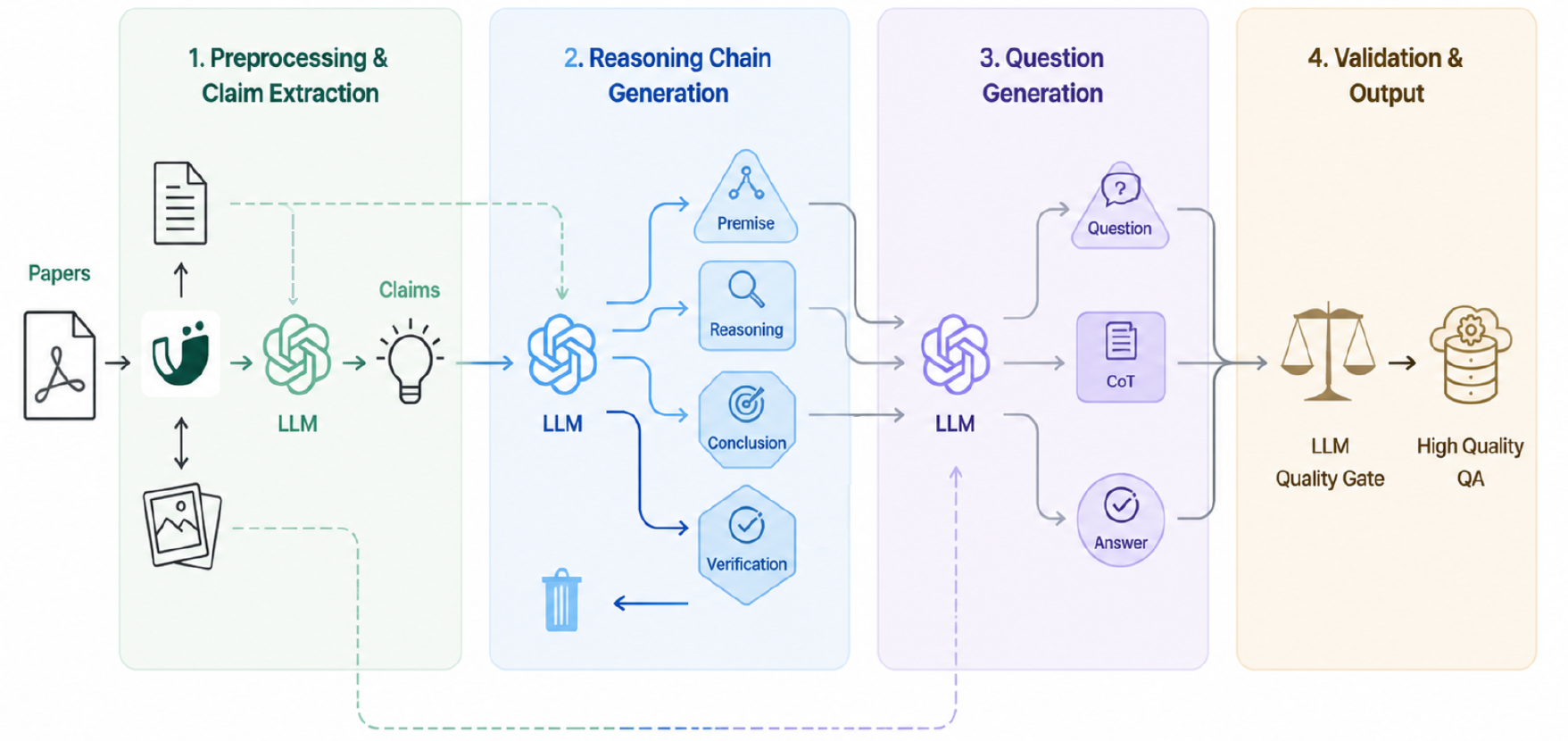}
  \caption{Overview of the four-stage MechReason construction pipeline: core claim extraction, reasoning-chain decomposition, anti-shortcut question generation, and quality verification. The pipeline separates logical reasoning from posterior corroboration evidence to construct multi-image, multi-hop mechanical reasoning questions.}
  \label{fig:pipeline}
\end{figure*}

The question generation pipeline of MechReason takes mechanical engineering papers as input and converts authentic scientific arguments into challenging multi-image, multi-hop QA samples. As shown in Fig.~\ref{fig:pipeline}, the pipeline consists of a preprocessing step followed by four stages: core claim extraction, reasoning-chain decomposition, anti-shortcut question generation, and quality verification.

\textbf{Preprocessing.}
For each paper, we first use MinerU \cite{wang2024mineru} to convert PDF documents into structured text and visual materials. The parsed results include full-text content, figure captions, surrounding figure contexts, and corresponding image files. We then apply post-processing scripts to inspect parsing quality and correct obvious structural errors, ensuring that downstream stages can reliably access both textual and visual evidence.

\textbf{Stage 1: Core Claim Extraction.}
The first stage identifies core mechanical engineering claims from the full paper. We use multimodal large language models to traverse the paper text, figure captions, and figure contexts, and select conclusions that lie near the logical endpoint of the authors' argumentation. A qualified claim should be mechanically meaningful, supported by multiple pieces of evidence, and require a non-trivial reasoning process rather than direct extraction from a single figure or table.

\textbf{Stage 2: Reasoning-Chain Decomposition.}
For each extracted claim, the second stage reconstructs its supporting argument and decomposes it into four components:
\begin{itemize}
    \item \textbf{Preconditions} contain directly accessible facts required to start the reasoning process, such as relevant images, geometric parameters, material properties, operating conditions, and boundary constraints. They should not include derived conclusions.
    \item \textbf{Derivation} describes how the conclusion is obtained from the preconditions, including physical laws, mathematical relations, causal mechanisms, and engineering trade-offs used in the paper.
    \item \textbf{Conclusion} corresponds to the core claim extracted in Stage 1 and serves as the endpoint of the reasoning chain.
    \item \textbf{Corroboration} records posterior validation evidence, such as experimental measurements, simulation results, or performance comparisons, that verifies the conclusion after it has been derived.
\end{itemize}

The central design in this stage is to distinguish the logical basis of a conclusion from the evidence used to validate it. Mechanical engineering papers often present intuitive figures, such as final performance comparisons or stress contour maps, to make complex conclusions easier to verify. If these materials are exposed in the question, models may answer by reading the validation result rather than reconstructing the causal chain. Therefore, Stage 2 explicitly marks such posterior materials as corroboration, while preserving the derivation as reasoning supervision.

\textbf{Stage 3: Anti-Shortcut Question Generation.}
The third stage transforms each decomposed reasoning unit into an evaluable multimodal QA sample. For each unit, we use the \texttt{conclusion} as the source of the target answer and compress it into a concise, deterministic short answer. The question is then generated around this conclusion as a clear engineering reasoning objective. To make the question solvable, we insert necessary information from the \texttt{preconditions}, including materials, structures, operating conditions, boundary conditions, or design constraints.

The \texttt{derivation} is not included in the question context. Instead, it is organized as chain-of-thought supervision that records the intermediate reasoning process from preconditions to conclusion. The \texttt{corroboration} component is excluded from the visible input to prevent shortcut solutions based on final experimental results, performance tables, or verification figures. In this way, the model must infer the answer from the provided premises and relevant visual evidence, rather than locate it from posterior validation materials.

\textbf{Stage 4: Quality Verification and Metadata Annotation.}
The fourth stage applies multimodal quality verification to candidate QA samples. The verification model receives the question, answer, reasoning process, and corresponding images, and checks whether each sample satisfies four criteria: prerequisite sufficiency, visual dependency, reasoning complexity, and mechanical relevance.

Prerequisite sufficiency requires that a solver with mechanical engineering knowledge can derive the answer from the question and provided images. Visual dependency ensures that the question genuinely relies on visual information. Reasoning complexity filters out samples that can be solved by direct table lookup, numerical reading, OCR, or single-step calculation. Mechanical relevance removes questions whose core reasoning is not grounded in mechanical engineering. Samples passing this quality gate are retained and annotated with metadata such as question type, reasoning category, visual materials, and reasoning supervision.

Overall, MechReason does not directly extract QA pairs from papers. It first identifies mechanically meaningful claims, reconstructs their argumentation chains, converts them into anti-shortcut questions, and filters low-quality samples through multimodal verification. This process keeps each sample grounded in the original scientific argument while requiring genuine multi-hop mechanical reasoning.
\section{Experiments}

\begin{table*}[t]
\centering
\caption{Main results on MechReason. Accuracy (\%) is reported for each reasoning subcategory and overall performance.}
\label{tab:mainResults}
\resizebox{\textwidth}{!}{
\begin{tabular}{lccccccccc}
\toprule
\multirow{2}{*}{Model}
& \multicolumn{2}{c}{Explanatory}
& \multicolumn{2}{c}{Predictive}
& \multicolumn{2}{c}{Design}
& \multicolumn{2}{c}{Diagnostic}
& \multirow{2}{*}{Overall} \\
\cmidrule(lr){2-3}
\cmidrule(lr){4-5}
\cmidrule(lr){6-7}
\cmidrule(lr){8-9}
& Mech. & Dev.
& Perf. & Proc.
& Entity & Workflow
& Limit & Failure
& \\
\midrule

% ----- Closed-source models -----
\midrule
\rowcolor{gray!10}\multicolumn{10}{c}{\textbf{Closed Source Models}} \\
\midrule
GPT-5.5              & \textbf{60.03} & \textbf{63.16} & \textbf{87.36} & \textbf{77.44} & \textbf{58.49} & \textbf{54.78} & \textbf{48.82} & 54.10          & \textbf{62.89} \\
Claude Opus-4.8     & 52.41          & 48.77          & 85.60          & 71.76          & 58.32          & 51.30          & 46.83          & \textbf{62.30} & 58.07          \\
Gemini-3.1-Pro      & 43.26          & 48.99          & 80.46          & 69.92          & 49.80          & 49.35          & 43.31          & 50.82          & 51.99          \\
Gemini-3.5-Flash    & 44.52          & 47.37          & 80.84          & 72.93          & 50.00          & 44.16          & 40.94          & 44.26          & 51.64          \\

% ----- Open-source models -----
\midrule
\rowcolor{gray!10}\multicolumn{10}{c}{\textbf{Open Source Models}} \\
\midrule
GLM-4.1-9B-thinking            & 35.48          & \textbf{40.08} & 65.90          & \textbf{57.89} & \textbf{40.00} & \textbf{41.13} & 31.50          & 34.43          & \textbf{42.11} \\
Kimi K2.6           & \textbf{36.61} & 36.03          & \textbf{66.28} & 51.13          & 35.71          & 39.83          & \textbf{33.86} & \textbf{37.70} & 40.83          \\
LLaVA-OneVision-1.5-8B & 26.93          & 28.34          & 63.22          & 36.84          & 35.10          & 34.63          & 24.41          & 34.43          & 34.47          \\
Qwen3.5-9B          & 24.54          & 28.34          & 61.30          & 48.87          & 33.13          & 29.00          & 17.32          & 18.03          & 32.35          \\
InternVL3.5-8B      & 24.12          & 23.08          & 65.13          & 43.61          & 30.61          & 28.14          & 13.39          & 14.75          & 30.84          \\
\bottomrule
\end{tabular}
}
\end{table*}

\subsection{Experimental Setup}

We evaluate a diverse set of representative multimodal large language models (MLLMs) on the MechReason test set, including both proprietary and open-source models. The proprietary models include GPT-5.5 \cite{singh2025openai}, Claude Opus-4.8 \cite{anthropic2026opus48}, Gemini-3.5-Flash \cite{google2026gemini35flash} and Gemini-3.1-Pro \cite{google2026gemini31pro}, while the open-source models include Kimi K2.6 \cite{team2026kimi}, Qwen3.5-9B \cite{qwen35blog}, GLM-4.1-9B-thinking \cite{vteam2025glm45vglm41vthinkingversatilemultimodal}, LLaVA-OneVision-1.5-8B-Base \cite{an2025llava}, and InternVL3.5-8B-Instruct \cite{wang2025internvl3}. All models are evaluated under the same protocol: the input consists of the question text and the corresponding multiple images, and the model is required to generate a short answer. During evaluation, no external tools, retrieval systems, or additional paper contexts are provided, ensuring a fair comparison across different models.

We adopt accuracy as the primary evaluation metric. Since answers in MechReason are typically concise engineering conclusions derived from scientific papers, model predictions may differ from the reference answers due to reasonable paraphrasing. Therefore, instead of exact string matching, we employ a strong judge model(Kimi K2.6) to perform semantic consistency evaluation. The judge only receives the question, reference answer, and final prediction, and does not access model identities. Specifically, for each sample $i$, we provide the question $q_i$, the reference answer $a_i$, and the predicted answer $\hat{a}_i$ to the judge model $J$, which is instructed to determine whether $\hat{a}_i$ is semantically consistent with $a_i$ under the given question context. The overall accuracy is defined as:

\begin{equation}
\mathrm{Acc}=\frac{1}{N}\sum_{i=1}^{N}\big[J(q_i,a_i,\hat{a}_i)=1\big],
\end{equation}

where $N$ denotes the number of test samples. For a specific reasoning category $c$, the category-wise accuracy is calculated as:

\begin{equation}
\mathrm{Acc}_c=\frac{1}{|D_c|}\sum_{i\in D_c}\big[J(q_i,a_i,\hat{a}_i)=1\big],
\end{equation}

where $D_c$ represents the subset of samples belonging to category $c$. This evaluation protocol preserves the deterministic nature of short-answer evaluation while allowing reasonable variations in expressions, making it more suitable for assessing conclusion-oriented question answering in mechanical engineering literature.

\subsection{Main Results}

Table~\ref{tab:mainResults} reports the performance of different MLLMs on MechReason. GPT-5.5 achieves the best overall accuracy of 62.89\%, followed by Claude Opus-4.8 with 58.07\%. Gemini-3.1-Pro and Gemini-3.5-Flash obtain similar performance, reaching 51.99\% and 51.64\%, respectively. Among open-weight models, GLM-4.1V performs best with 42.11\%, while LLaVA-OneVision-1.5, Qwen3.5-9B, and InternVL3.5-8B obtain 34.47\%, 32.35\%, and 30.84\%, respectively. These results show a clear gap between the strongest proprietary models and current open-weight MLLMs. Nevertheless, even the best model still fails on more than one third of the test samples, indicating that complex multi-image mechanical reasoning remains far from solved.

The results also reveal substantial differences across reasoning types. Predictive reasoning is consistently easier than the other categories: most models achieve their highest scores on performance prediction and process extrapolation, where curve trends, tabulated conditions, or explicit comparisons among configurations often provide relatively direct evidence. In contrast, explanatory, design, and diagnostic reasoning remain more challenging. These tasks require models to integrate evidence across multiple images and infer mechanisms, feasible engineering choices, or failure boundaries, rather than simply extract visible facts. The low performance of open-weight models on limit diagnosis and failure diagnosis further suggests that current MLLMs still struggle with boundary conditions, physical constraints, and model-assumption failures.

\subsection{Failure Mode Analysis}

\begin{table*}[t]
\centering
\caption{Distribution of primary trace-aware error types among incorrect responses. Percentages are computed over the incorrect samples of each model.}
\label{tab:errorTypes}
\begin{tabular*}{\textwidth}{l@{\extracolsep{\fill}}c*{5}{S[table-format=2.2]}}
\toprule
Model & N &
{\makecell{Grounding}} &
{\makecell{Chain}} &
{\makecell{Assumption}} &
{\makecell{Hallucination}} &
{\makecell{Compression}} \\
\midrule
GPT-5.5          & 839  & 21.78 & 37.12 & 10.43 & 2.87 & 27.80 \\
Gemini-3.1-Pro   & 1085 & 25.74 & 44.65 & 5.35  & 1.01 & 23.25 \\
Kimi K2.6        & 1339 & 33.21 & 42.79 & 8.85  & 6.15 & 19.00 \\
Qwen3.5-9B       & 1530 & 44.86 & 37.23 & 6.98  & 5.93 & 5.00  \\
\bottomrule
\end{tabular*}
\end{table*}

We further conduct a trace-aware failure analysis on incorrect responses from GPT-5.5, Gemini-3.1-Pro, Kimi K2.6, and Qwen3.5-9B. Following prior studies on multi-hop evidence grounding, reasoning shortcuts, false-premise robustness, hallucination, and chain-of-thought faithfulness~\cite{yang2018hotpotqa,jiang2019avoiding,yu2023crepe,turpin2023language}, we classify each observable response into five error types: \textit{Grounding}, where the response misreads or fails to align with key visual/textual evidence; \textit{Chain}, where relevant evidence is partially recognized but connected through an incorrect multi-hop reasoning path; \textit{Assumption}, where the response relies on an invalid condition, boundary, or applicability range; \textit{Hallucination}, where unsupported facts or premises are introduced; and \textit{Compression}, where the visible reasoning is partially aligned but the final short answer loses the intended conclusion.

As shown in Table~\ref{tab:errorTypes}, the dominant failure source is \textbf{Chain} error, accounting for 37.12\%--44.65\% of failures across all models. This indicates that models often identify some relevant evidence but fail to compose it into the correct mechanical mechanism or engineering conclusion. \textbf{Grounding} errors are especially prominent for Qwen3.5-9B, reaching 44.86\%, suggesting that weaker models still struggle with locating and aligning heterogeneous visual evidence. In contrast, stronger models show larger proportions of \textbf{Compression} errors, such as GPT-5.5 at 27.80\% and Gemini-3.1-Pro at 23.25\%, where the response trace is partially reasonable but the final answer is too generic or misses a key qualifier. Hallucination remains relatively rare, showing that MechReason is challenging less because models invent arbitrary content, and more because they must ground evidence, preserve assumptions, and complete multi-hop mechanical reasoning chains.

%We further conduct output-level failure attribution on incorrect responses from GPT-5.5, Gemini-3.1-Pro, Kimi K2.6, and Qwen3.5-9B. A discriminative judge compares the question, reference answer, and model prediction, and assigns each error to one dominant category based only on the observable final answer.

%As shown in Table~\ref{tab:errorTypes}, errors are highly concentrated in \textbf{Target Misalignment} and \textbf{Mechanism Misinterpretation}, which together account for 92.43\% of all failures. Qwen3.5-9B shows the strongest target misalignment, suggesting that weaker models often produce context-related but conclusion-misaligned answers. In contrast, stronger models exhibit more mechanism-level errors, indicating that even when the target is recognized, models may still fail to recover the correct causal pathway or structure--performance relation. Other error types remain minor, showing that MechReason is mainly challenging because models must align heterogeneous evidence with the intended engineering claim and reconstruct the underlying mechanical mechanism, rather than merely avoid hallucination or recognize isolated visual details.

\subsection{Chain-of-Thought Supervision Analysis}

\begin{figure}[t]
  \centering
  \includegraphics[width=\columnwidth]{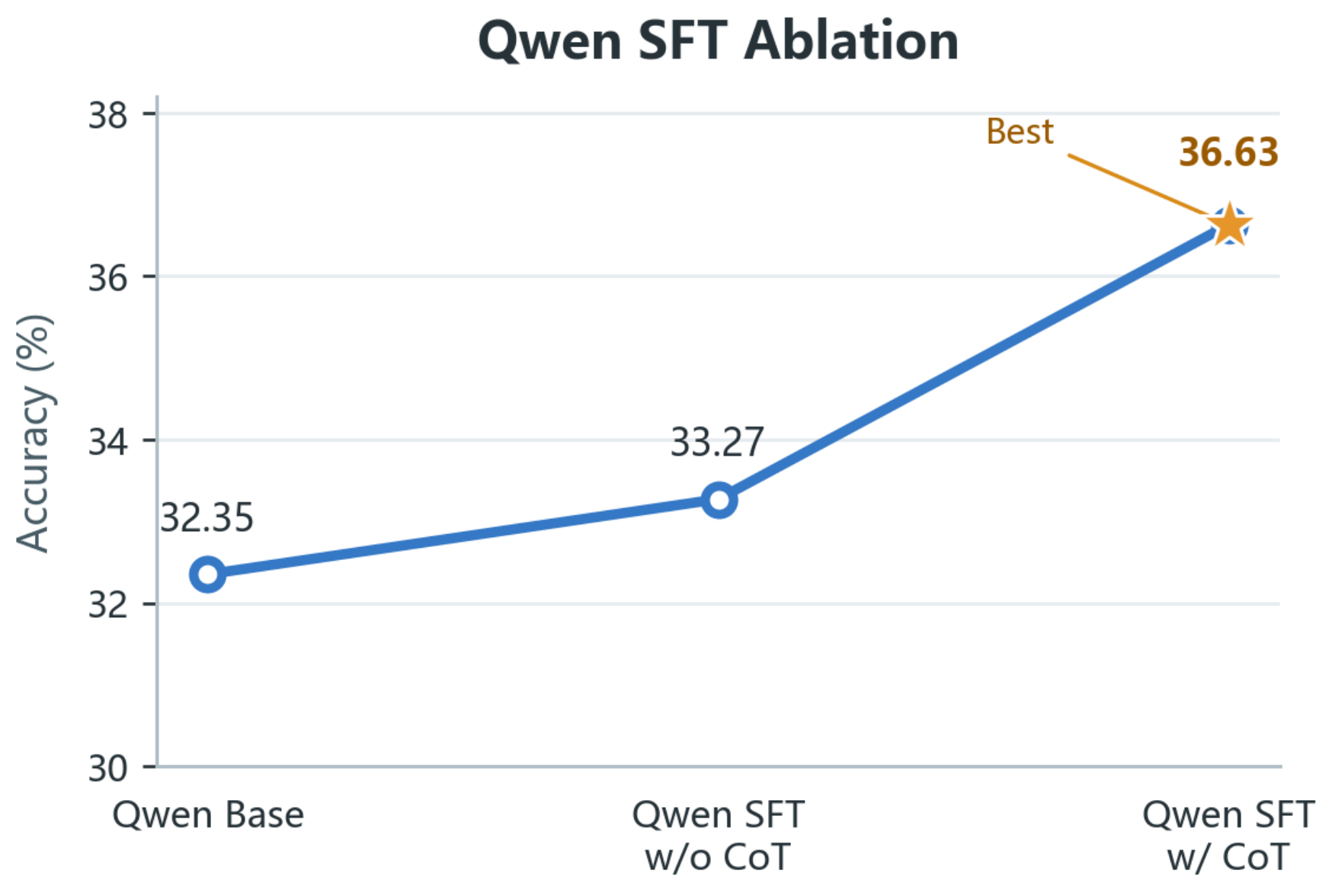}
  \caption{Overall accuracy of Qwen3.5-9B before and after supervised fine-tuning with and without Chain-of-Thought (CoT) supervision.}
  \label{fig:sft}
\end{figure}

To investigate the effect of reasoning-chain supervision, we compare three settings: the original Qwen3.5-9B model, a version fine-tuned with only short-answer supervision, and a version fine-tuned with explicit chain-of-thought (CoT) supervision. As shown in Fig.~\ref{fig:sft}, fine-tuning without CoT only improves accuracy from 32.35\% to 33.27\%, yielding a marginal gain of 0.91 percentage points. In contrast, CoT-based fine-tuning increases accuracy to 36.63\%, improving over the original model by 4.27 points and over non-CoT fine-tuning by 3.36 points. This suggests that the improvement does not mainly come from adapting answer formats or domain terminology, but from learning intermediate engineering reasoning processes.

\begin{table}[t]
\centering
\caption{Subcategory-wise sample counts and accuracy changes after CoT supervision. \(\Delta\) denotes the percentage-point change over the original Qwen3.5-9B model.}
\label{tab:subcatChange}
\begin{tabular}{l
                 S[table-format=3.0]
                 S[table-format=+2.2]}
\toprule
{Subcategory} & {N} & {\(\Delta\)} \\
\midrule
Mechanism Analysis       & 713 & +2.95 \\
Deviation Attribution    & 247 & -0.81 \\
Performance Prediction   & 261 & +8.43 \\
Process Extrapolation    & 133 & +10.53 \\
Entity Design            & 490 & +6.46 \\
Workflow Planning        & 231 & +3.03 \\
Limit Diagnosis          & 127 & +0.00 \\
Failure Diagnosis        & 61  & +4.92 \\
\bottomrule
\end{tabular}
\end{table}

Table~\ref{tab:subcatChange} shows that CoT supervision brings uneven gains across reasoning subcategories. The largest improvements appear in \textbf{Process Extrapolation}, \textbf{Performance Prediction}, and \textbf{Entity Design}, with gains of 10.53, 8.43, and 6.46 points, respectively. These tasks usually have clearer condition--response or option--performance structures, where explicit reasoning traces can directly teach the model how to connect parameter changes, operating conditions, design constraints, and final engineering consequences.

In contrast, \textbf{Limit Diagnosis} shows no improvement, and \textbf{Deviation Attribution} slightly decreases. These tasks often require precise localization of visual evidence, recognition of boundary conditions or modeling assumptions, and attribution of errors to material, process, or measurement factors. Such capabilities are difficult to acquire from textual CoT supervision alone. Overall, CoT supervision improves reasoning patterns with explicit derivation paths, but it does not eliminate the broader bottlenecks in multi-image evidence grounding and mechanical mechanism understanding.
\section{Conclusion}
We introduce MechReason, a multi-image, multi-hop benchmark for complex mechanical engineering reasoning. To our knowledge, it is the first benchmark of this kind constructed from real mechanical engineering papers, which provide complex engineering claims, explicit reasoning chains, and heterogeneous visual evidence. MechReason covers four reasoning intentions, including Explanation, Prediction, Design, and Diagnosis, and contains approximately 12K QA pairs with CoT supervision across nine types of mechanical visual evidence.

Experiments show that current MLLMs remain far from reliable on MechReason: even GPT-5.5 achieves only 62.89\% accuracy. Error analysis indicates that failures are dominated by chain errors, grounding errors, and compression errors, rather than simple hallucination. CoT supervision improves tasks with clearer derivation paths, but brings limited gains for diagnosis and deviation attribution. These results suggest that future models need stronger abilities to integrate multi-image evidence, recover mechanical mechanisms, and reason under engineering constraints.
\bibliography{aaai2027}


\end{document}